\documentclass[pdflatex,sn-mathphys-num]{sn-jnl}

\usepackage{graphicx}%
\usepackage{multirow}%
\usepackage{amsmath,amssymb,amsfonts}%
\usepackage{amsthm}%
\usepackage{mathrsfs}%
\usepackage[title]{appendix}%
\usepackage{xcolor}%
\usepackage{textcomp}%
\usepackage{manyfoot}%
\usepackage{booktabs}%
\usepackage{algorithm}%
\usepackage{algorithmicx}%
\usepackage{algpseudocode}%
\usepackage{listings}%
\usepackage{pifont}

\theoremstyle{thmstyleone}%
\theoremstyle{thmstyletwo}%

\theoremstyle{thmstylethree}%

\begin{document}

\title[Article Title]{ProtoFlow: Interpretable and Robust Surgical Workflow Modeling with Learned Dynamic Scene Graph Prototypes}


\author*[1,2,3]{\fnm{Felix} \sur{Holm}}\email{felix.holm@tum.de}

\author[2]{\fnm{Ghazal} \sur{Ghazaei}}

\author[1,3]{\fnm{Nassir} \sur{Navab}}

\affil[1]{\orgdiv{Chair for Computer-Aided Medical Procedures}, \orgname{Technical University Munich}, \orgaddress{\city{Munich}, \country{Germany}}}

\affil[2]{\orgdiv{Corporate Research and Technology}, \orgname{Carl Zeiss AG}, \orgaddress{\city{Munich}, \country{Germany}}}

\affil[3]{\orgname{Munich Center for Machine Learning (MCML)}, \orgaddress{\city{Munich}, \country{Germany}}}




\abstract{\textbf{Purpose:}  
Detailed surgical recognition is critical for advancing AI-assisted surgery, yet progress is hampered by high annotation costs, data scarcity, and a lack of interpretable models. While scene graphs offer a structured abstraction of surgical events, their full potential remains untapped. In this work, we introduce ProtoFlow, a novel framework that learns dynamic scene graph prototypes to model complex surgical workflows in an interpretable and robust manner.

\textbf{Methods:} ProtoFlow leverages a graph neural network (GNN) encoder-decoder architecture that combines self-supervised pretraining for rich representation learning with a prototype-based fine-tuning stage. This process discovers and refines core prototypes that encapsulate recurring, clinically meaningful patterns of surgical interaction, forming an explainable foundation for workflow analysis.

\textbf{Results:} We evaluate our approach on the fine-grained CAT-SG dataset. ProtoFlow not only outperforms standard GNN baselines in overall accuracy but also demonstrates exceptional robustness in limited-data, few-shot scenarios, maintaining strong performance when trained on as few as one surgical video. 
Our qualitative analyses further show that the learned prototypes successfully identify distinct surgical sub-techniques and provide clear, interpretable insights into workflow deviations and rare complications.

\textbf{Conclusion:} By uniting robust representation learning with inherent explainability, ProtoFlow represents a significant step toward developing more transparent, reliable, and data-efficient AI systems, accelerating their potential for clinical adoption in surgical training, real-time decision support, and workflow optimization.
}




\maketitle

\section{Introduction}\label{sec1}

The development of AI-driven surgical automation and decision support systems relies heavily on accurate recognition and understanding of complex surgical workflows. However, surgical video analysis presents unique challenges, including significant variability across procedures, real-time constraints, and the necessity for structured yet interpretable representations of surgical activities. Traditional deep learning-based methods primarily focus on frame-wise classification tasks such as phase recognition \cite{Twinanda2016EndoNet,Jin2018ToolNet,watchandlearn}, tool detection \cite{Sahu2020,Funke2019}, or semantic scene segmentation \cite{Shvets2018}. While effective for isolated recognition tasks, these methods struggle to capture the intricate and evolving interdependent relationships between surgical elements, limiting both \textbf{model generalizability and interpretability}, two key factors for deploying AI-driven decision support in clinical environments.

A promising alternative lies in \textbf{scene graphs}, structured graphical representations that explicitly model objects and their interactions within a scene \cite{Krishna2017,Johnson2015Image}. \textbf{Dynamic Scene Graphs (DSGs)} extend this concept by capturing the temporal evolution of relationships in complex domains such as surgical workflow~\cite{holm2023dynamic}. 
In surgical workflow modeling, recent works have leveraged scene graphs for surgical workflow recognition in laparoscopic cholecystectomy \cite{murali2023encoding}, simulated knee replacement surgery \cite{oezsoy2022_4D_OR} and cataract surgery \cite{holm2023dynamic,koksal2024sangria} using positional and temporal relations between the surgical elements. Despite these advances, existing approaches still face important hurdles, primarily due to three key limitations:  
(1) \textbf{Data scarcity.} Acquiring large-scale, expert-labeled surgical datasets is costly and time-consuming, limiting the applicability of supervised learning approaches.  
(2) \textbf{Generalization challenges.} Surgical workflows exhibit high variability across patients, techniques, and clinical settings, making it difficult for models to generalize effectively. Edge cases and anomalous events, in particular, are the hardest to model yet often the most critical to handle accurately.
(3) \textbf{Interpretability and Transparency.} Furthermore, there is a pressing need for improved interpretability, so that medical teams can understand and trust the decisions made by AI-based systems.

To address these challenges, we introduce \textbf{ProtoFlow}, a novel \textbf{prototype learning framework} for surgical workflow modeling, inspired by recent advances in prototype learning for graph-based representations \cite{Dai25Prot}. Instead of treating each surgical scene as an independent instance, ProtoFlow identifies and learns recurring structured representations of surgical interactions as \textbf{workflow prototypes}. These prototypes encapsulate fundamental patterns in surgical workflows, enabling a more interpretable and generalizable representation of procedure dynamics. By leveraging \textbf{graph autoencoder architectures}, our approach learns \textbf{abstract prototype embeddings}, which serve as structured class representations that enhance both accuracy and interpretability in surgical workflow modeling.

A key advantage of \textbf{ProtoFlow} is its ability to perform \textbf{few-shot learning}, significantly reducing the reliance on large labeled datasets. By integrating \textbf{self-supervised and supervised learning}, we enhance prototype robustness, enabling effective generalization across different surgical procedures, even in low-data scenarios.
We evaluate our approach on \textbf{cataract surgery}, utilizing a newly introduced \textbf{CAT-SG} scene graph dataset \cite{CATSG} featuring positional, geometrical, temporal and semantic relationships among surgical elements. Cataract surgery is a highly standardized yet delicate procedure, with \textbf{significant intraoperative variability} influenced by factors such as cataract grade, patient-specific conditions, and surgical technique adaptations. 

Beyond surgical workflow recognition, ProtoFlow supports \textbf{qualitative deviation analysis}, providing interpretable insights into deviations from standard surgical procedure. While prior work has explored \textbf{deviation detection} in surgical workflows, such as in laparoscopic rectopexy \cite{RectopexyDeviations2022}, these approaches lack explicit interpretability and structured workflow representations. By integrating prototype learning and scene graphs, our method enhances explainability, making it potentially useful for surgical training, workflow optimization, and AI-driven decision support. 

Our contributions are summarized as follows:
\textbf{Architecture:} We propose ProtoFlow, a novel prototype learning framework for surgical workflow modeling using dynamic scene graphs, integrating supervised and self-supervised learning.
\textbf{Comprehensive Evaluation:} We conduct ablation studies and benchmarking on the CAT-SG dataset~\cite{CATSG}, demonstrating that ProtoFlow outperforms existing baselines. Our few-shot learning experiments highlight the framework’s ability to generalize more effectively on as few as one surgical video.
\textbf{Enhanced Interpretability:} We explore qualitative deviation analysis and explainability, leveraging prototype-driven representations to provide interpretable insights into surgical workflow variations and model decisions. Critically, we find ProtoFlow handles low-frequency phases and anomalous events with greater reliability.

\section{Methodology}

\begin{figure}
    \centering
    \includegraphics[width=\linewidth]{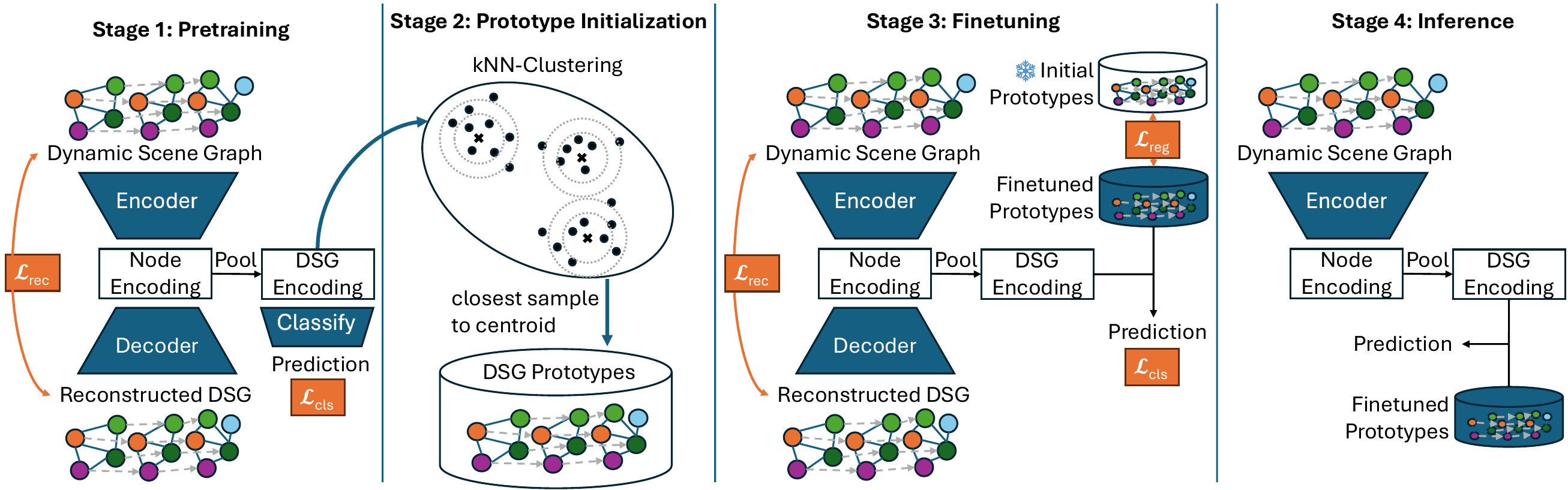}
    \caption{Overview of the four-stage pipeline (pretraining, prototype initialization, fine-tuning, and inference) for prototype-based surgical workflow modeling (ProtoFlow).}
    \label{fig:workflow}
\end{figure}

Our goal is to learn interpretable, prototype-based representations of dynamic scene graphs for surgical workflow modeling. Our pipeline, therefore, predicts the surgical phase directly from dynamic scene graphs. The pipeline proceeds in four stages: (1) Pretraining a graph autoencoder model for graph reconstruction, (2) Prototype initialization, (3) Fine-tuning with prototype-based supervision, and (4) Inference. An overview is shown in Figure~\ref{fig:workflow}.

\bmhead{GNN Architecture and Pretraining}

Let each DSG at time $t$  be denoted by  $G_{t} = (A_{t}, X_{t})$, where  $A_{t} \in \mathbb{R}^{N \times N}$  is the adjacency matrix (encoding spatial, temporal, and semantic relationships among the  $N$  nodes), and  $X_{t} \in \mathbb{R}^{N \times d}$  is the node feature matrix with $d$ features per node. Our model uses a GNN encoder  $f_{\theta}$  that transforms each graph into node embeddings $ H_{t}\in \mathbb{R}^{N \times d_h}$ with embedding dimension $d_h$.
A global DSG-level embedding $z_{t} \in \mathbb{R}^{d_h}$  is then obtained by pooling  $H_{t}$ (e.g., via mean pooling).

To stabilize the latent space and promote discriminative embeddings, we use a decoder $g_{\phi}$ to reconstruct the original graph. Given  $H_{t}$, the decoder outputs reconstructed node features $\hat{X}_{t}$ and (optionally) adjacency $\hat{A}_{t}$. The reconstruction loss is defined as a mean-squared error (MSE) loss.

Additionally, a classification head  $h_{\psi}$  takes  $z_{t}$  and predicts a surgical workflow phase $y_{t}$. We adopt a cross-entropy classification loss for phase prediction.
The overall pretraining objective is $
    \mathcal{L}_{\text{pretrain}} \;=\; \mathcal{L}_{\text{rec}} \;+\; \,\mathcal{L}_{\text{cls}}
$.

\bmhead{Prototype Initialization}

After pretraining, we use the trained encoder  $f_{\theta}$  to map all training DSGs to embeddings  $z_{t}$. For each class  $c \in \{1,\dots,C\}$, we cluster the embeddings that belong to class $c$ into $K$ groups (using k-nearest neighbors). Each cluster’s centroid is taken as an initial prototype. These prototypes serve as reference points in the embedding space for each surgical phase.
\begin{equation}
  \mathcal{P} \;=\; \{\mathbf{p}_{c,k} \mid c=1,\ldots,C;\; k=1,\ldots,K\}.  
\end{equation}

\bmhead{Fine-tuning with Prototype Regularization}

In fine-tuning, both the GNN encoder $f_{\theta}$  and the prototypes $\mathbf{p}_{c,k}$ are updated to optimize classification based on distance to each prototype. Given an input DSG embedding  $z_{t}$, we compute the distance to each prototype, softmax the distance vector for normalisation and then average the softmaxed distances per class to convert to prediction $\hat{y}_{t,c}$:
\begin{equation}
d_{c,k}(z_{t}) = \lVert z_{t} - \mathbf{p}_{c,k}\rVert,
\quad
\hat{y}_{t,c} \;=\; \frac{1}{K} \sum_{k=1}^{K} \frac{\exp\bigl(-d_{c,k}(z_t)\bigr)}{\displaystyle \sum_{c{\prime}=1}^{C}\sum_{k{\prime}=1}^{K}\exp\bigl(-d_{c{\prime},k{\prime}}(z_t)\bigr)}.
\end{equation}

We then apply a cross-entropy loss on $\hat{y}_{t}$, denoted $\mathcal{L}_{\text{cls}}$.
To preserve the cluster structure that emerged in pretraining, we regularize prototype movement with a MSE term:
\begin{equation}
  \mathcal{L}_{\text{reg}} = \sum_{c,k} \lVert \mathbf{p}_{c,k}^{\,(0)} - \mathbf{p}_{c,k}\rVert^{2},
\end{equation}

where $\mathbf{p}_{c,k}^{\,(0)}$ are the initial prototypes from clustering. We also maintain the graph reconstruction loss $\mathcal{L}_{\text{rec}}$. Altogether, $
    \mathcal{L}_{\text{finetune}} \;=\; \mathcal{L}_{\text{cls}} \;+\; \mathcal{L}_{\text{reg}} \;+\; \mathcal{L}_{\text{rec}}
$.

\bmhead{Inference and Explainability}

At inference, we encode a new DSG, $G_{\ast}$  to  $z_{\ast}$, measure its Euclidean distances to each prototype, and produce a softmax-based prediction. Because each prototype $\mathbf{p}_{c,k}$ is tied to representative substructures of a surgical phase, these distance values provide intuitive explanations, indicating which prototypical scenario most closely matches the new graph. Further interpretability arises from comparing node-level or edge-level alignments between the input graph and the closest prototypes.

\section{Experiments}
We conduct a series of experiments to evaluate the performance of our proposed ProtoFlow approach for surgical workflow recognition on the CAT-SG dataset. Our experiments are designed to (1) compare ProtoFlow against existing baselines, (2) analyze the impact of different temporal windows within the Dynamic Scene Graph Prototypes, (3) highlight generalizability via few-shot learning experiments, and (4) showcase the interpretability of our learned prototypes.

\bmhead{CAT-SG Dataset}
We use the recently introduced CAT-SG dataset, which extends the original CATARACTS dataset~\cite{CATARACTS} with semantic scene graphs. Nodes in each scene graph represent surgical objects (e.g., instruments, anatomical structures), while edges capture a variety of relationships such as spatial proximity and semantic interactions. In total, the dataset comprises 50 cataract surgery videos, each ranging from 6 to 40 minutes in length. The videos in the dataset are segmented into 19 different surgical steps, describing the workflow of the cararact surgery in high detail. The dataset is split into training, validation, and test sets at the video level to ensure no overlap of procedures across folds. We follow the proposed split of CATARACTS, yielding 25 training, 5 validation, and 20 test videos~\cite{CATARACTS}.

To leverage the graphs of the CAT-SG temporally, we construct dynamic scene graphs, similar to~\cite{holm2023dynamic}. Scene graphs for multiple frames are aggregated by connecting nodes of the same class from temporally adjacent frames with temporal edges, resulting in one large graph.

\bmhead{Experimental Setup: Backbone GNN}
We implement the encoder as a multi-layer GNN using 3 GATv2~\cite{gatv2} layers, and the decoder as a 3-layer MLP. The hidden dimension for both models is 1024, the size of the graph encoding $z_{t}$ is set to 512.\\
\textbf{Prototype Module.} We initialize  19 (\#classes) $\times$ 3  prototypes (\#prototypes per class) from cluster centroids in the latent space after pretraining. 
\\
\textbf{Optimization.} We train all methods for a maximum of 50 (pretraining) + 50 (finetuning)  epochs with an initial learning rate of $3 \times 10^{-4}$ and a stepped learning rate schedule using the Adam optimizer with a batch size of 64. Early stopping based on validation set performance is used to prevent overfitting. We use pytorch~\cite{paszke2019pytorch} and pytorch geometric~\cite{Fey2019FastGR} as our framework and train our models on NVIDIA A40 GPUs with 40GB VRAM, taking approximately 24 hours per run.\\ 
\textbf{Temporal Window Setup.}
We explore different temporal window sizes (1~frame to 60~frames, corresponding to up to 180 seconds) to measure the effect of temporal context on accuracy and F1 scores (see Tab.~\ref{tab:comparison} and Fig.~\ref{fig:temp}).\\
\textbf{Evaluation Metrics.}
We report Accuracy and F1 Score as standard metrics for workflow recognition. These metrics are averaged over the videos, to give each video equal weight in the final result. Since our model is initialized from scratch, we always show the average results from three runs for our results and indicate the standard deviation.

\section{Results \& Discussion}



\begin{figure}
    \centering
    \includegraphics[width=0.7\linewidth]{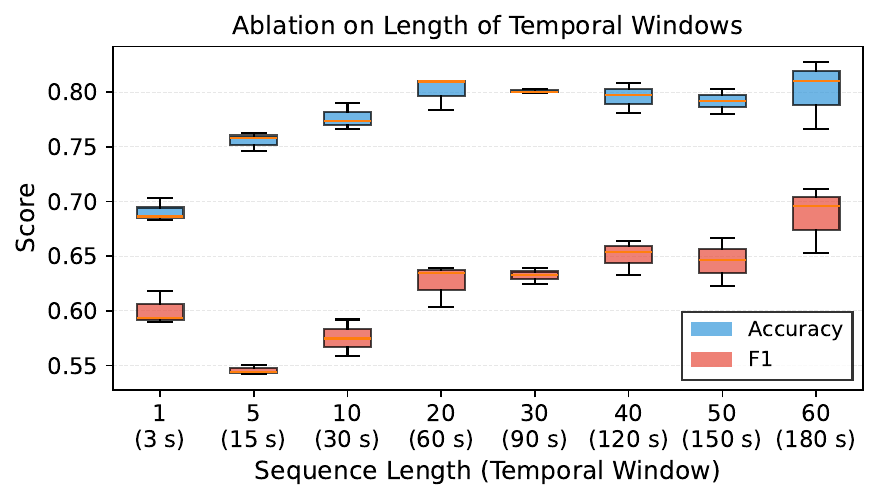}
    \caption{Impact of temporal window size on ProtoFlow's performance (Accuracy in blue, F1 in red). The model's performance improves as the temporal context increases, but the gains start to plateau after 20-30 frames (60-90s). This indicates diminishing returns and suggests that a context of roughly one minute is sufficient for effective workflow recognition.}
    \label{fig:temp}
\end{figure}

\bmhead{Comparison with Baselines for Surgical Workflow Recognition} Table~\ref{tab:comparison} presents the workflow recognition results of our method alongside baselines. 
We compare ProtoFlow to baselines previously reported in \cite{CATSG}, including GNN- and LLM-based approaches, all predicting surgical phase from (dynamic) scene graphs. 
When employing only one frame (i.e., no temporal aggregation), ProtoFlow already achieves comparable performance.
However, as we expand the temporal window from 1~frame to 30–60~frames (see Fig.~\ref{fig:temp}), ProtoFlow’s performance markedly improves, surpassing 80\% in Accuracy and reaching 68.66\% in F1 score at 60 frames.
Our results confirm that capturing the evolving dependencies and temporal context strengthens the latent representations for surgical workflow recognition. Notably, using more than 20~frames (60s) yields good results but strongly diminishing returns on the increased graph complexity, suggesting that modeling roughly one minute of context is generally sufficient.\\
\begin{table}[b]
\caption{Surgical workflow recognition results on CAT-SG \cite{CATSG}}\label{tab:comparison}
\begin{tabular}{@{}lcccc@{}}
\toprule
& Temporal & Semantic & & \\
 Method         &  Window&  Relations&  Accuracy& F1 \\ 
\cmidrule(r){1-1} \cmidrule(lr){2-3} \cmidrule(lr){4-5}
Holm et al. \cite{holm2023dynamic} &  1 frame &  \ding{55} CATARACTS &  65.56 & 52.24\\
         GATv2 \cite{CATSG} & 1 frame & \ding{51} CAT-SG & 70.81 & 56.02 \\
         Llama 3.2 3B \cite{CATSG}  & 1 frame & \ding{51} CAT-SG & 69.13 & 53.87 \\ 
         Holm et al. \cite{holm2023dynamic} & 30 frames (90 s) & \ding{55} CATARACTS & 73.77 & 64.93 \\
         GATv2 \cite{CATSG} & 30 frames (90 s) & \ding{51} CAT-SG & 78.63 & \textbf{70.15} \\ \midrule
        ProtoFlow & 1 frame& \ding{51} CAT-SG& 69.05 $\pm$ 1.07& 60.06 $\pm$ 1.55\\ 
         ProtoFlow & 30 frames (90 s)& \ding{51} CAT-SG &  \underline{80.07} $\pm$ 0.19& 63.21 $\pm$ 0.72\\
         ProtoFlow & 60 frames (180 s)& \ding{51} CAT-SG &  \textbf{80.14} $\pm$ 3.15& \underline{68.66} $\pm$ 3.06\\
\botrule
\end{tabular}
\end{table}
%
\bmhead{Robustness in Limited-Data Scenarios}
Gathering surgery data remains expensive and time-consuming. We thus assessed ProtoFlow in few-shot settings, where only a handful of fully labeled surgical videos (1, 2, or 5) were available for training. As shown in Table~\ref{tab:results_fewshot}, our prototype-driven approach significantly outperforms the GATv2 baseline, with a margin of 19 up to 24 percentage points (p.p.) in Accuracy. Even when only a single procedure was used for training, ProtoFlow achieved over 39\% Accuracy (versus \~15\% for GATv2), highlighting how self-supervised prototype learning confers stronger generalization than purely supervised GNN methods. These findings underscore the potential value of our method in real-world clinical scenarios, where annotated data for specific procedure variations can be scarce. 
\begin{table}[b]
\caption{Few-shot surgical workflow recognition with varying number of annotated videos}
    \begin{tabular}{@{}lccc@{}}
        \toprule
         Method &   \#~Annotated Videos &  Accuracy&  F1\\ 
         \cmidrule(r){1-1}\cmidrule(lr){2-2}\cmidrule(lr){3-4}
        GATv2&   \multirow{2}{*}{1}&  15.43 $\pm$ 0.12 &  9.49 $\pm$ 0.07\\
         ProtoFlow &  &  \textbf{39.49} $\pm$ 2.27 &  \textbf{28.89} $\pm$ 2.57\\ \midrule
         GATv2&  \multirow{2}{*}{2}&  40.65 $\pm$ 1.03 &  30.17 $\pm$ 2.43\\
         ProtoFlow &  &  \textbf{58.63} $\pm$ 2.15 &  \textbf{41.11} $\pm$ 2.49\\ \midrule
         GATv2&   \multirow{2}{*}{5}&  43.56 $\pm$ 2.87&  38.14 $\pm$ 3.27\\
         ProtoFlow &  &  \textbf{62.66} $\pm$ 4.82 &  \textbf{44.03} $\pm$ 6.1\\ \bottomrule
    \end{tabular}
    \label{tab:results_fewshot}
\end{table}

\begin{figure}
    \centering
    \includegraphics[width=1\linewidth]{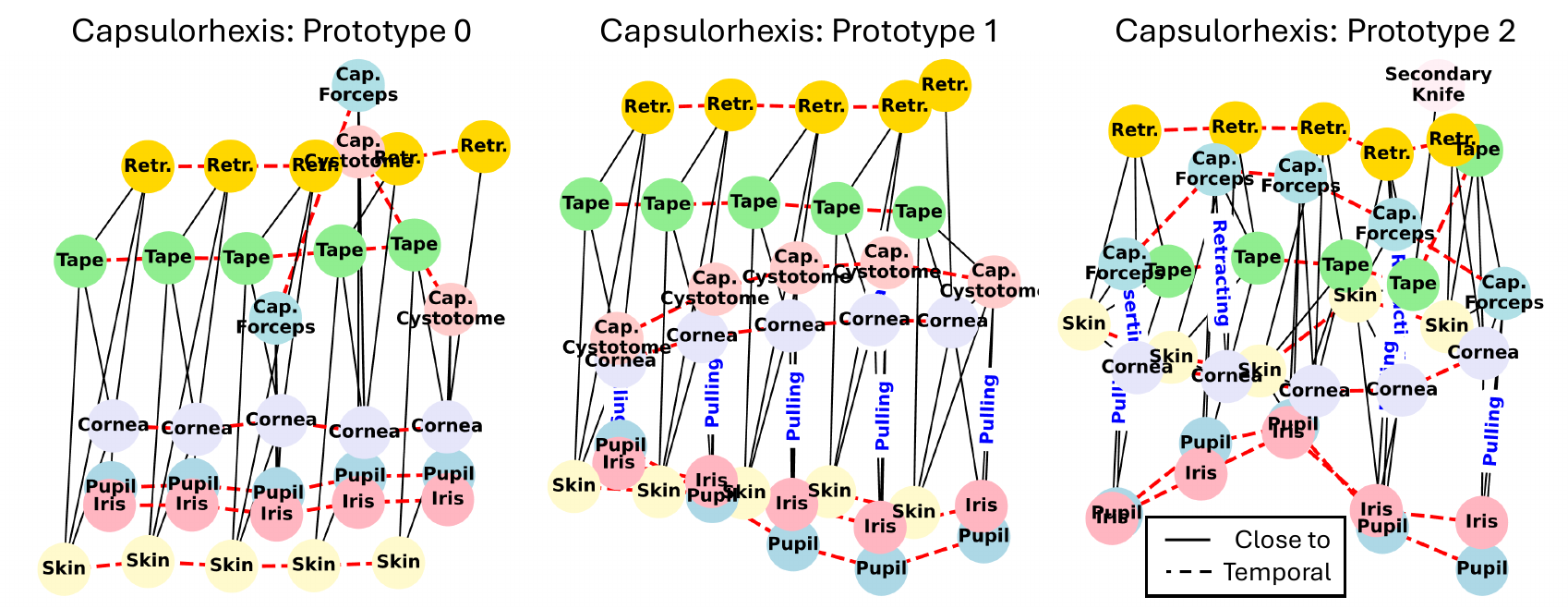}
    \caption{Learned prototypes (temporal window = 5 frames) for Capsulorhexis, each capturing a distinct instrument usage pattern discovered by the model (use of cystotome, forceps, or both).}
    \label{fig:rhexis}
\end{figure}

\begin{figure}
    \centering
    \includegraphics[width=1\linewidth]{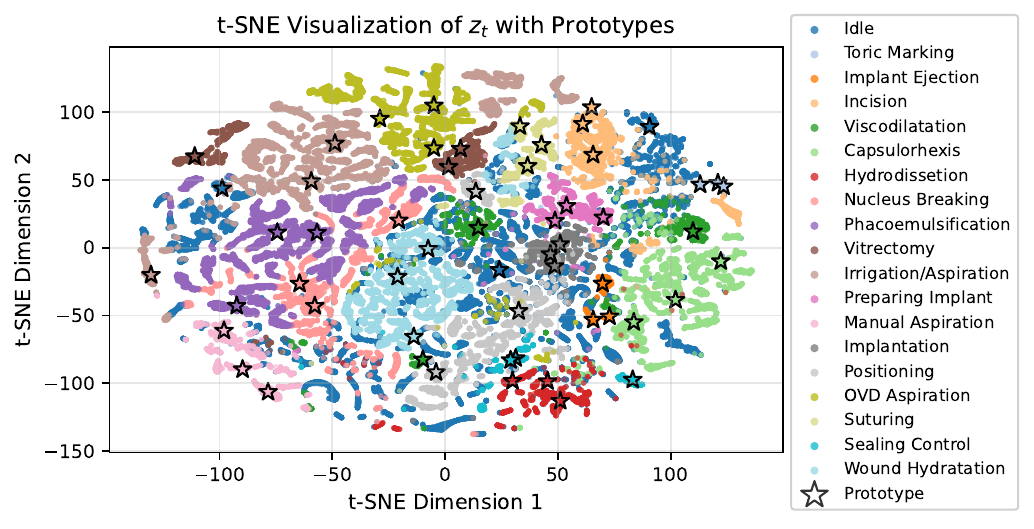}
    \caption{t-SNE visualization of the learned graph embeddings ($z_t$), with each color representing a different surgical phase. The black stars indicate the positions of the prototype embeddings. The prototypes are situated within their corresponding class clusters, effectively capturing distinct variations in surgical action patterns. This is particularly evident for the "Capsulorhexis" phase (bottom right), where the embeddings visibly separate into sub-groups that align with the different instrument-use patterns discovered by the model (as shown in Fig.~\ref{fig:rhexis}), highlighting its ability to automatically uncover meaningful workflow variations.}
    \label{fig:tsne}
\end{figure}

%
%
\bmhead{Learned Prototype Groups} A key advantage of ProtoFlow lies in its self-explainable structure, derived from prototype embeddings that capture recurring scene-graph patterns. Fig.~\ref{fig:rhexis} illustrates three learned prototypes for the “Capsulorhexis” phase, each corresponding to a distinct surgical variation: some surgeons employ only a cystotome, others prefer forceps, and a third group uses both instruments in tandem. Notably, our pipeline discovers these variations entirely via clustering on the latent embeddings, without requiring explicit labeling of which instrument technique was used. Such interpretability aids surgeons and data scientists in verifying whether the model’s “understanding” of a phase aligns with known, clinically meaningful sub-techniques. It could also promote establishing more standardized workflows or verify adherence to predefined ones. The t-SNE visualization of the embedding space in Fig.~\ref{fig:tsne} further illustrates this automatic discovery. The samples for the "Capsulorhexis" phase, located in the bottom right of the plot, visibly separate into roughly three distinct sub-clusters. These groupings correspond directly to the different surgical sub-techniques captured by the prototypes in Fig.~\ref{fig:rhexis}. This provides strong visual confirmation that our method automatically identifies and separates these clinically meaningful variations within a single surgical phase, without any explicit supervision. The prototypes (starred points) are spread throughout the clusters of each class, showing that they successfully represent distinct variations in action patterns for each phase of the workflow.

\begin{figure}
    \centering
    \includegraphics[width=1\linewidth]{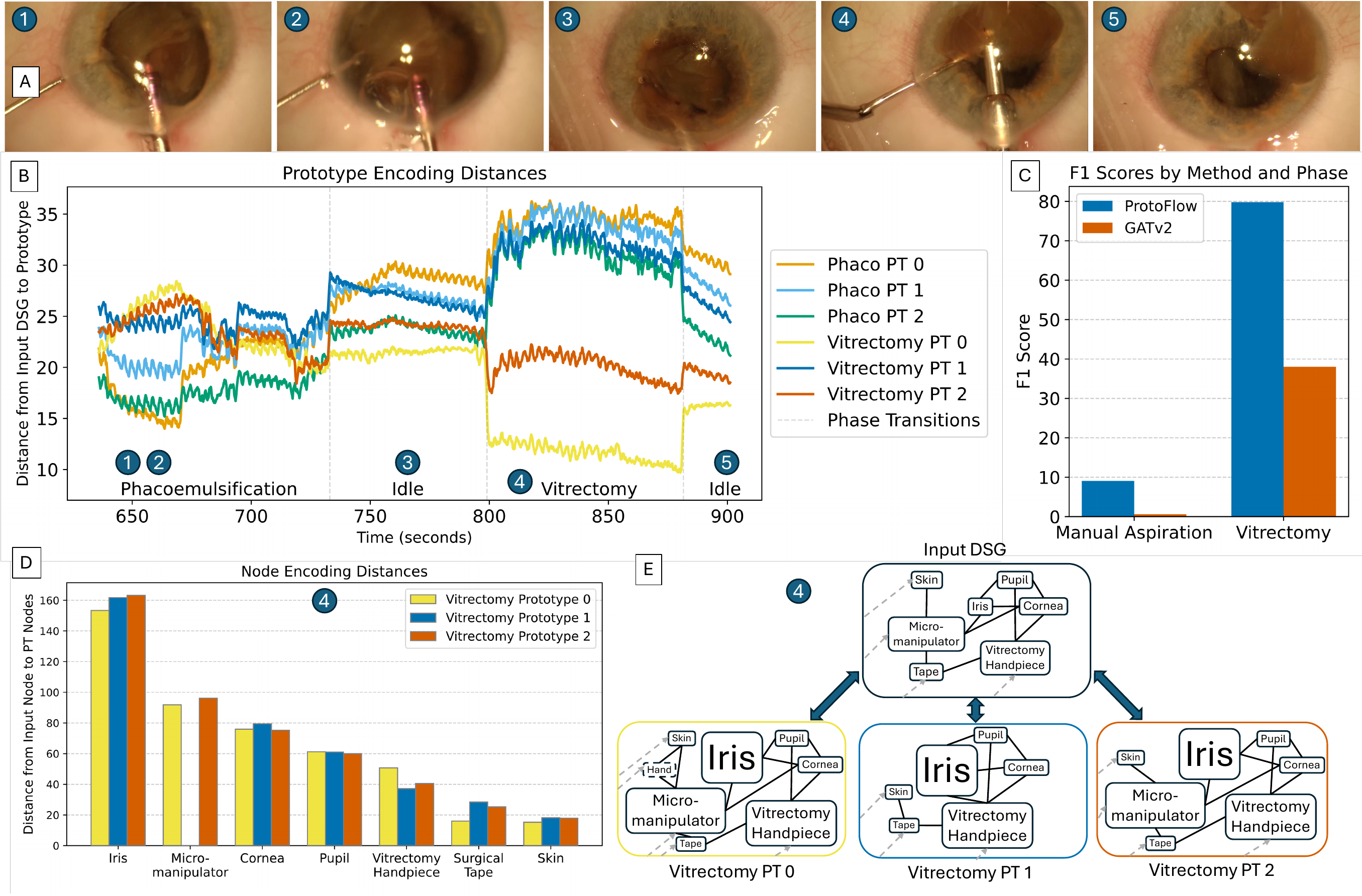}
    \caption{ProtoFlow's interpretability in handling a rare phase (Vitrectomy) with a complication (prolapsed iris). (A) Frames from the test video show the unexpected event. (B) As the procedure transitions from Phacoemulsification to Vitrectomy (~800s), the distance to the Phacoemulsification prototypes spikes, flagging a workflow deviation. (C) Quantitative results show ProtoFlow robustly identifies the rare phase. (D, E) The model's interpretability allows for a granular explanation: the iris node in the input graph is a major outlier compared to the learned Vitrectomy prototypes, providing a quantitative signal directly corresponding to the visible complication.
    A video of this sequence is attached in the Supplementary.}
    \label{fig:robust}
\end{figure}

\bmhead{Handling Workflow Deviations} Beyond recognizing standard workflows, ProtoFlow demonstrates resilience in handling low-frequency or emergent surgical phases. Fig.~\ref{fig:robust} highlights this capability using a challenging sequence from a test video where a rare Vitrectomy phase occurs—an event seen only once during training. This situation is further complicated by the presence of a prolapsed iris, a rare surgical complication (Fig.~\ref{fig:robust}A).
As the surgeon transitions from the Phacoemulsification phase to the unexpected Vitrectomy, the model immediately registers a significant deviation. As shown in Fig.~\ref{fig:robust}B, the distance from the live scene graph to all learned Phacoemulsification prototypes sharply increases. This spike acts as an automatic flag, signaling that the current workflow is no longer consistent with the established patterns of the preceding phase. Despite the event's rarity, ProtoFlow correctly classifies the phase as Vitrectomy, demonstrating robust performance where baseline models fail (Fig.~\ref{fig:robust}C).
The framework's interpretability, however, extends beyond just flagging a phase change. By matching node encodings between the input DSG and the learned prototypes, we can generate a more granular explanation. The analysis reveals that the iris node in the input graph is a significant outlier when compared to the corresponding nodes in the standard Vitrectomy prototypes (Fig.~\ref{fig:robust}D). This highlights an anomalous state in the iris's representation, providing a clear, quantitative signal that directly corresponds to the observed clinical complication (Fig.~\ref{fig:robust}E).
This level of interpretability is crucial when encountering unexpected or emergent events; the system not only flags the anomaly but also provides evidence allowing an expert to trace the signal back to the specific components of the scene that triggered its detection. This capability to provide transparent, multi-level explanations for its reasoning in complex situations is essential for building trust and facilitating the clinical adoption of AI systems.


\bmhead{End-to-End Results}
While our main experiments use ground truth (GT) Dynamic Scene Graphs, we also validate our method's practical performance in an end-to-end setting. We trained and tested our model using graphs generated by CatSGG+~\cite{CATSG}, the current SOTA scene graph generation model for CAT-SG ~\cite{CATSG}. The results in Table~\ref{tab:results_e2e} indicate a minimal performance gap between using generated graphs and GT, demonstrating our method's robustness to prediction noise. This robustness is promising, as it paves the way for using self- or weakly-supervised scene graph generators to reduce the significant effort of manual annotation.


\begin{table}[b]
\caption{End-to-end surgical workflow recognition results}
    \begin{tabular}{@{}lcc@{}}
        \toprule
         Method &  Accuracy&  F1\\ 
         \cmidrule(r){1-1}\cmidrule(lr){2-3}
        
        GT + ProtoFlow&  \textbf{80.07} $\pm$ 0.19& \textbf{63.21} $\pm$ 0.72\\
         CatSGG+\cite{CATSG} + ProtoFlow&  78.66 $\pm$ 1.27 &  60.43 $\pm$ 0.87\\
         \bottomrule
    \end{tabular}
    \label{tab:results_e2e}
\end{table}




\section{Conclusion \& Future Work}

This work establishes that learning dynamic scene graph prototypes is a highly effective strategy for surgical workflow modeling, directly addressing the critical barriers of data scarcity and model opacity. We demonstrated that ProtoFlow achieves competitive accuracy while delivering exceptional robustness in limited-data, few-shot scenarios , significantly lowering obstacles for development in data-scarce clinical settings.
Beyond data efficiency, ProtoFlow's interpretability provides a tangible path toward the transparent AI systems required for clinical adoption. The model automatically discovers clinically meaningful workflow variations, like Capsulorhexis sub-techniques, offering a verifiable foundation for analysis. This capability extends to complex, rare events: the model robustly identifies an unexpected Vitrectomy phase and provides granular, node-level explanations for complications like an iris prolapse. Such detailed, multi-level explanations are essential for building clinical trust.
By uniting prototype-based interpretability with robust representation learning , ProtoFlow takes an essential step toward more explainable, reliable, and efficient AI solutions for next-generation surgical data science.

\bmhead{Supplementary information} We attach a video of the analysis shown in Figure~\ref{fig:robust}.
\bmhead{Acknowledgements} The authors thank Carl Zeiss AG for supporting this work.
\bmhead{Declarations} The authors have no competing interests to declare that are relevant to the content of this article.

\bibliography{references}

\end{document}